\documentclass[review]{elsarticle}
\usepackage{times}
\usepackage{soul}
\usepackage{url}
\usepackage[hidelinks]{hyperref}
\usepackage[utf8]{inputenc}
\usepackage[small]{caption}
\usepackage{graphicx}
\usepackage{amsmath}
\usepackage{amsthm}
\usepackage{booktabs}
\usepackage{algorithm}
\usepackage{algorithmic}
\usepackage{multirow}
\usepackage{mathtools}
\usepackage{amsfonts} 
\usepackage{color}
\usepackage{epsfig}
\usepackage{amssymb}
\usepackage{eso-pic}
\usepackage{xspace}
\usepackage{comment}

\makeatletter
\DeclareRobustCommand\onedot{\futurelet\@let@token\@onedot}
\def\@onedot{\ifx\@let@token.\else.\null\fi\xspace}

\makeatother

\usepackage{amssymb}
\usepackage{lineno,hyperref}

\journal{Journal of \LaTeX\ Templates}

\begin{document}

\begin{frontmatter}

\title{Unsupervised Video Summarization with a Convolutional Attentive Adversarial Network}

\author[NWPU]{Guoqiang Liang\fnref{mycorrespondingauthor}}

\fntext[mycorrespondingauthor]{Corresponding author: Guoqiang Liang}
\ead{gqliang@nwpu.edu.cn}

\author[NWPU]{Yanbing Lv}
\author[NWPU]{Shucheng Li}
\author[NWPU]{Shizhou Zhang}
\author[NWPU]{Yanning Zhang}
% full name
\address[NWPU]{National Engineering Laboratory for Integrated Aero-Space-Ground-Ocean Big Data Application
Technology, School of Computer Science, Northwestern Polytechnical University, Xi'an, China}

\begin{abstract}
With the explosive growth of video data, video summarization, which attempts to seek the minimum subset of frames while still conveying the main story, has become one of the hottest topics. Nowadays, substantial achievements have been made by supervised learning techniques, especially after the emergence of deep learning. However, it is extremely expensive and difficult to collect human annotation for large-scale video datasets. To address this problem, we propose a convolutional attentive adversarial network (CAAN), whose key idea is to build a deep summarizer in an unsupervised way. Upon the generative adversarial network, our overall framework consists of a generator and a discriminator. The former predicts importance scores for all frames of a video while the latter tries to distinguish the score-weighted frame features from original frame features. Specifically, the generator employs a fully convolutional sequence network to extract global representation of a video, and an attention-based network to output normalized importance scores. To learn the parameters, our objective function is composed of three loss functions, which can guide the frame-level importance score prediction collaboratively. To validate this proposed method, we have conducted extensive experiments on two public benchmarks SumMe and TVSum. The results show the superiority of our proposed method against other state-of-the-art unsupervised approaches. Our method even outperforms some published supervised approaches.
\end{abstract}


\end{frontmatter}
%\linenumbers

\section{Introduction}
With the increasing popularity and decreasing cost of video capture devices, there has been a phenomenal surge in videos captured everyday. Facing with the sheer amount of video data, it is becoming more and more difficult for a user to browse all these videos completely to seek useful information. As a result, it has been very demanding to develop some intelligent techniques for efficiently retrieving and analyzing this large amount of videos. 

Video summarization \cite{li2020exploring,2015Video}, one of the most promising techniques, aims to condense a video into a brief summary. Compared with original video, the summary should preserve its main semantic information while largely shorten the length. Currently, there are two types of resulted summary: (i) a set of selected key-frames, (ii) a set of selected video segments. The former is a set of frames while the latter is a set of temporally continuous segments including key-shots. In this paper, we mainly discuss the latter since it is more vivid. Like traditional methods, the critical process is to predict an importance score for every frame in a video, which is then used to generate the key-shots.

One of the challenges in video summarization is how to learn the complex temporal relationship of a video. Due to the great success and wide application of deep learning \cite{he2016deep}, video summarization has been formulated as a sequence labeling problem, where recurrent neural networks (RNN) are used to model temporal dependencies \cite{KeZhang2016,Ji2020video}. Many existing methods \cite{Ji2020video,Mrigank2018,Jiri2018} try to solve the video summarization problem in a supervised way, where the model is learned by minimizing the distance between the predicted summary and the ground-truth summary annotated by human. Benefiting from some well-annotated datasets, they have achieved good performance. However, it is tedious and costly to obtain frame-level or shot-level labels for a large number of videos. Therefore, some researchers move to unsupervised methods. Based on Generative Adversarial Network (GAN) \cite{Goodfellow2014}, Mahasseni et al. \cite{Mahasseni2017} designed an unsupervised summarization framework consisting of a summarizer and a discriminator, both of which are based on long short memory network (LSTM). The summarizer first selects some key frames and then reconstructs the video. The discriminator aims at distinguishing between the original video and its reconstruction. Based on this, more variants are developed to improve the performance \cite{LYuan2019,Jung2019}. For example, Yuan et al. \cite{LYuan2019} proposed to perform a bi-directional reconstruction between the produced summary and its original video. Because of its low efficiency in modeling long-range relationship \cite{Joe2015,SVen2015}, some works are proposed to augment LSTM with attention module. He et al. \cite{He2019} combined the self-attention module and the LSTM in their generator. Similarly, Jung et al. \cite{jungglobal} designed a global and local decomposition strategy. However, the above methods still suffer from the high time complexity and difficulty in model parallelization due to the recurrent structure of LSTM.

In this paper, we propose a novel convolutional attentive adversarial network for unsupervised video summarization. Under the generative adversarial networks, our model includes a generator and a discriminator, which are jointly trained by minimizing the representation error between raw frame features and score-weighted frame features. To capture the various temporal dependence, our generator consists of a fully convolutional sequence network and a self-attention module. The former acts as a feature transformation to extract global representation while the self-attention mechanism can capture the long-range dependencies of a video. Finally, the generator outputs normalized importance scores. Since there is no available label information, we design a LSTM-based discriminator to guide the learning. Intuitively, the importance score represents the importance of a frame for retaining the main semantic information, so we can multiply the original frame feature with the importance score to obtain a representation for the summary. By distinguish the raw frame features from importance score weighted frame features, the discriminator could provide the signal for training. Specifically, it should consider the original feature as true and the score-weighted feature as false. 

%The generator are jointly trained so as to maximally confuse the discriminator LSTM and make generated weighted frame features sufficiently similar to the raw frame features.
To train the model's parameters, our loss function contains three parts. The first is the traditional adversarial loss to make the generator against with the discriminator. Besides the adversarial loss, we also employ sparsity loss and the reconstruction loss. The former is to limit the total number of selected key frames as in \cite{Mahasseni2017}. The latter is to keep the content of a summary similar to its original video. Once converged, the generator can produce reasonable frame-level importance scores. During testing, we will discard the discriminator and only use the generator. Since there is no recurrent processing units, our generator can be paralleled much easier compared with other LSTM-based methods. To validate the proposed method, we have conducted extensive experiments on the SumMe \cite{Gygli2014} and TVSum \cite{Yale2015}, which are two popular benchmarks for video summarization. The experimental results show the effectiveness of the method.

Compared with previous work, our main contribution includes: 1) We combine the self-attention mechanism with a fully convolutional sequence network to better capture the long-range temporal dependencies of videos; 2) We propose a convolutional attentive generative adversarial network for unsupervised video summarization; 3) On two popular benchmarks, i.e. TVSum and SumMe, our proposed unsupervised approach not only achieve better performance within unsupervised methods, but also is superior to most published supervised approaches.
%method is an instance of GAN with three losses. We propose a simple network based convolutional attentive GAN framework to directly minimize the distance between generated weighted frame features and the original appearance feature.the discriminator and the generator all consist of one multi-head self-attention module and one BiLSTM model

\section{Related Work}
In this section, we will review some related works on: i) supervised video summarization; ii) unsupervised video summarization; and iii) attention mechanism. Instead of early method based on handcrafted criteria \cite{liu2002optimization, Potapov2014}, current methods are dominated by deep learning. Therefore, we will focus on video summarization techniques based on deep learning.
\subsection{Supervised Video Summarization}
Supervised summarization methods use frame-level/shot-level annotations to train a model to predict the importance scores. Since they can utilize the human annotated ground truth, they tend to outperform unsupervised ones. Zhang et al. \cite{KeZhang2016} proposed a bidirectional LSTM network augmented with the Determinantal Point Process to increase the diversity of a summary. After this milestone work, many LSTM based methods are proposed \cite{Ji2020video,wei2018,wang2019stacked}. Ji et al. \cite{Ji2020video} devised a deep attentive encoder-decoder framework based on LSTM. In \cite{wang2019stacked}, memory layers are integrated with LSTM to further model long-term temporal dependency. Besides the LSTM architecture, some other modules are also used. For example, the fully convolution network original for image segmentation is adapted for video summarizaiotn \cite{Mrigank2018}. Park et al. \cite{park2019video} proposed to learn the relationships between action and scene. In \cite{BZhao2018}, Zhao et al. integrated shot segmentation and video summarization into a hierarchical structure-Adaptive RNN,

Benefiting the human annotated datasets, these methods have obtained excellent performance. However, due to the complexity of annotating the importance score of every frame in a video accurately, it is highly expensive to construct a large and various dataset. 
\subsection{Unsupervised Video Summarization} 
To address the above issues, many unsupervised methods are also proposed. By designing some heuristic criteria to make the summary to satisfy certain properties, they do not need human annotated importance score any more. In \cite{ZhouKaiyang2017}, video summarization was formulated as a deep reinforcement learning task with diversity and representativeness rewards. To alleviate its convergence problem, Chen et al. \cite{Chen2019WeaklySV} proposed a weakly supervised hierarchical reinforcement learning framework, which decompose the whole task into several subtasks.  Besides the reinforcement learning, GAN is also a universal framework for unsupervised learning. \cite{Mahasseni2017} firstly proposed a GAN based method, which includes three LSTM networks. Based on this, Jung et al. \cite{Jung2019} design a two-stream network which utilizes local and global temporal view. \cite{jungglobal} further combined the relative position embedding with this global and local decomposition strategy. Based on the successful cycle-GAN architecture in image generation, a  Cycle-consistent Adversarial LSTM architecture is proposed in \cite{LYuan2019}. In \cite{He2019}, a conditional feature selector is employed to provide conditional information for GAN. 

Compared with these approaches, we removes the recurrent structure, which is much easier to be paralleled. We only use the fully convolutional sequence network and self-attention module to capture global and local relationship.

\subsection{Attention Mechanism}
Similar to the visual attention of human beings, the attention mechanism in deep learning is to select information that is more critical to the current task from a lot of information. Considering the complex relationship of videos, the attention mechanism plays an important role in video summarization. In the early stages, some methods \cite{Ejaz2013,Yufei2005} calculated attention from low-level features, such as motion cue and salience map. However, these methods do not take the temporal dependencies among frames into consideration. In contrast, attention mechanism for sequences is very popular in natural language processing \cite{Bahdanau2014,Vaswan2017}. For example, Vaswani et al. \cite{Vaswan2017} first introduced the multi-head self attention mechanism, which allows to build relationship for long sequence. Without recurrent structure, it is very fast to complete the calculation with matrix multiplication. Due to its excellent effect, this mechanism becomes more and more popular in video summarizaiton. Recently, \cite{Ji2020video} used attention mechanism to mimic the way of selecting the key-shots of human. Fajtl et al. \cite{Jiri2018} put forward a self-attention network to generate the importance score of all frames. This powerful modeul is also used in \cite{He2019,Jung2019}. In this work, we combine the self-attention module with a GAN in a simple yet effective framework.

\section{Proposed method}
\begin{figure*}[t!]
	\centering
	\includegraphics[width=0.99\textwidth]{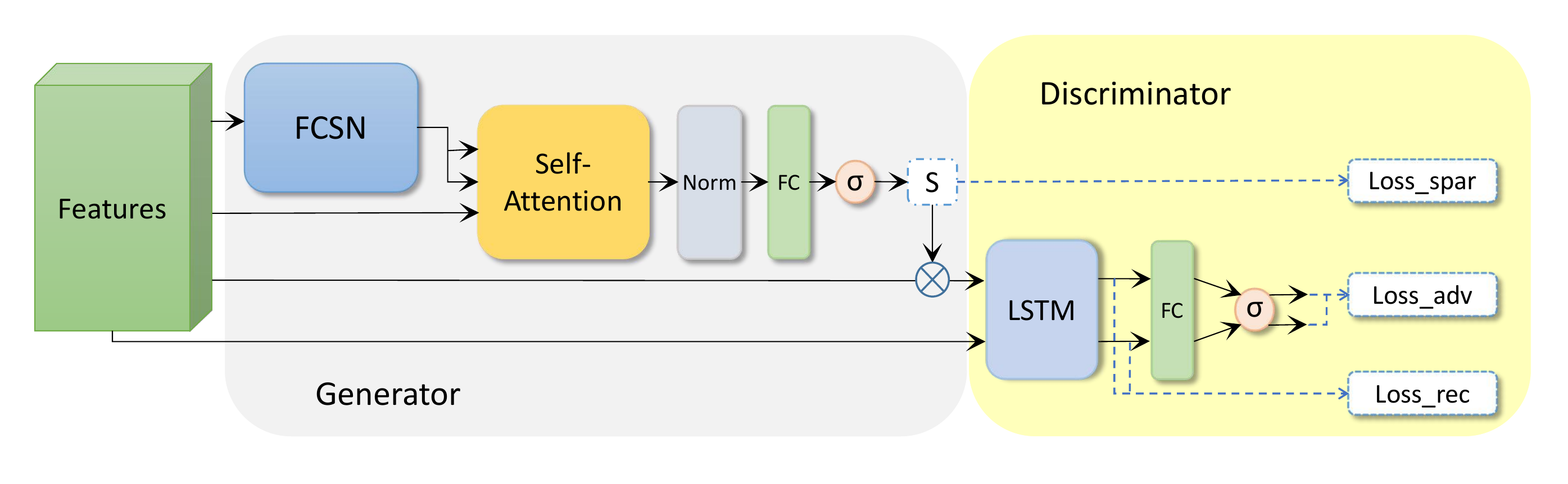}
	\caption{ The network architecture of our proposed approach, which mainly consists of two parts: a convolutional-attention generator (left) and a LSTM-based discriminator (right). Given a video sequence with $F$ frames, where each frame has an appearance feature $X_f$, the fully convolutional sequence network dynamically capture global information of all frames by several temporal pooling, convolution and deconvolution operations. Then an attention-based network is used to predict the frame-level importance score and generate weighted frame features. The discriminator based on a LSTM module is to differentiate the weighted frame feature from the original feature.}
	\label{Fig_frame}
\end{figure*}

This paper aims to address the unsupervised video summarization by jointly training a convolutional-attention generator and a discriminator in an adversarial manner. The overall architecture is illustrated in Fig. \ref{Fig_frame}. The generator consists of a fully convolutional sequence network and an attention-based network. Given a video $V={\left\{v_f\right\}}_{f=1}^F$ including $F$ frames in total, we first use a CNN to extract its appearance feature $X={\left\{x_f \in {\mathbb R}^d \right\}}_{f=1}^F$. Then, the fully convolutional sequence network takes this $X$ as input and generates an intermediate feature sequence $Y={\left\{y_f \in {\mathbb R}^d \right\}}_{f=1}^F$ which captures the global information of a video. The resulted sequence $Y$ and original feature $X$ are then fed together into the attention-based network, which produces a score sequence $S={\left\{s_f\in{\left[0,1\right]}\right\}}_{f=1}^F$ indicating the importance score of each frame.

Since there are no human-annotated labels, we design a discriminator consisting of a LSTM layer. By distinguishing original feature $X$ from score-weighted feature $\widetilde X$, it can provide the learning signal. The $\widetilde X$ is the score-weighted appearance feature, i.e. the element-wise product of generated scores $\left(s_1,\cdots,s_F\right)$ and the appearance features $\left(x_1,\cdots,x_F\right)$. The discriminator should classify them to two distinct classes: ‘original’ and ‘weighted’. In the following, we will detail the architecture of our generator and discriminator.
\subsection{Generator}
The temporal dependence plays a significant role in judging the importance of a frame. Moreover, it will be highly different for videos describing different events. To model this various dependence, the generator employs a fully convolutional sequence network \cite{Mrigank2018} to extract global representation of a video. Then an attention-based network is used to capture long range dependence, which finally produces frame-level importance scores.

\subsubsection{\textbf{Fully Convolutional Sequence Network}}
To encode the various temporal dependence of a video, we employ the fully convolutional sequence network (FCSN) as a global representation extractor, which performs convolution, deconvolution and pooling operations in the temporal dimension. Different from the original architecture \cite{Mrigank2018}, we add more skip connections between the convolution operations and the deconvolution operations. By fusing feature in different temporal ranges, the final feature will contain more information. Note that we currently deliver the whole video into the FCSN considering the length of videos in current datasets. In the future, we will design the hierarchical structure or decomposition strategy for very long videos.

\begin{figure*}[t!]
	\centering
	\includegraphics[width=0.95\textwidth]{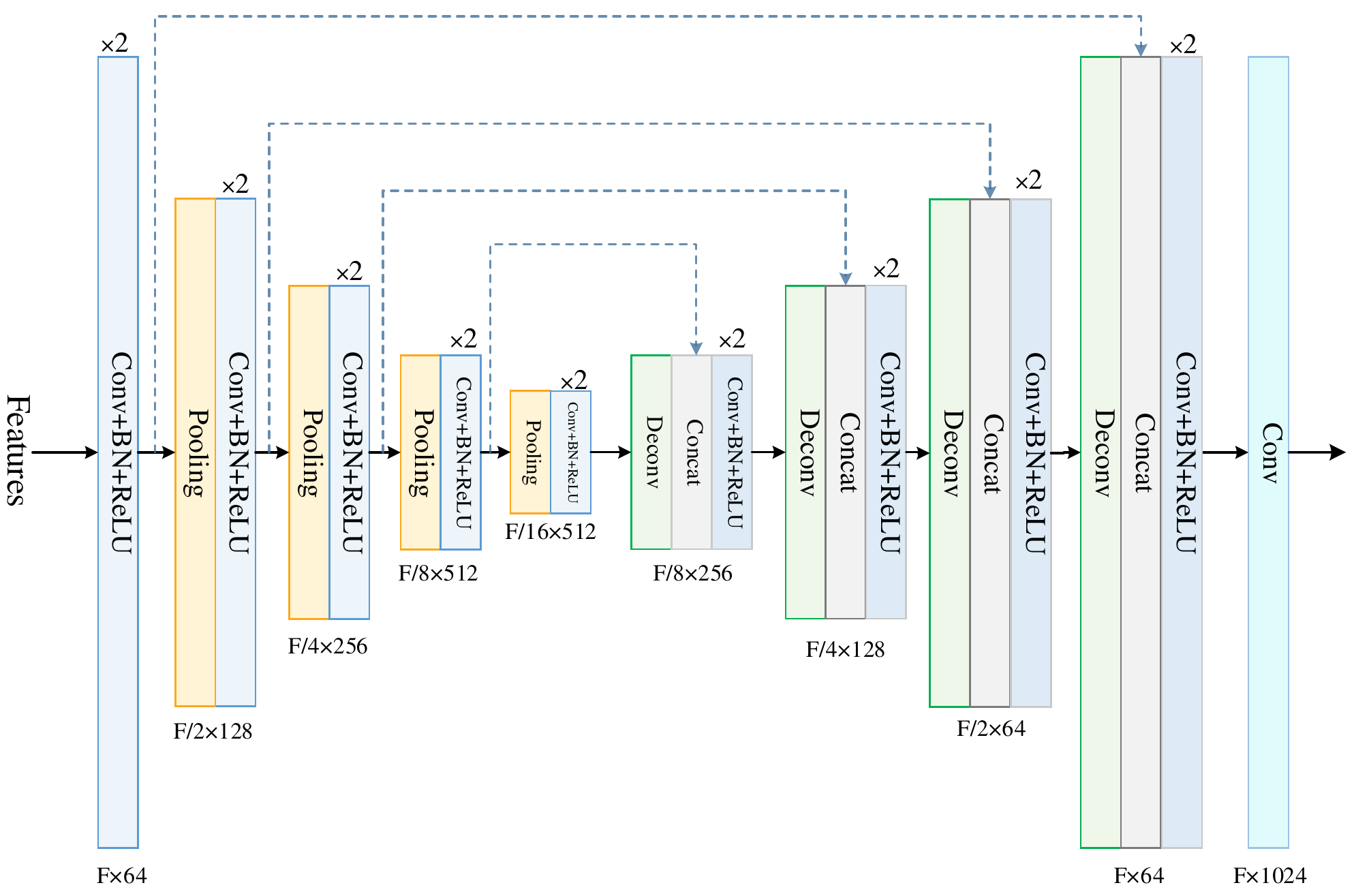}
	\caption{ The network architecture of the fully convolutional sequence network (FCSN).}
	\label{Fig_fcsn}
\end{figure*}

In detail, a $3 \times 3$ double-convolutional layer is firstly used to reduce the channel dimension of the frame feature to 64. A double-convolution layer is composed of two $3 \times 3$ temporal convolution layers, each of which is followed by a batch normalization and a ReLU activation. Then there are four temporal max-pooling layers per followed by a double-convolution layer. Each of these max-pooling layers reduces the temporal dimension by half while a double-convolution layer will double the channel dimension. To convert the temporal dimension of the feature to $F$, we adopt four temporal deconvolution operations along the time axis and four concatenation operations along the channel dimension. The concatenation operation merges the deconvolution feature map with the output of previous convolutional layer. After each concatenation, we employ a double-convolution layer to produce the desired output channel. The detailed temporal and channel dimension of resulted feature are shown in the Fig. \ref{Fig_fcsn}. Lastly, we apply a $1 \times 1$ convolution operation again and obtain a $d$ dimension feature for each frame.

\subsubsection{\textbf{Attention-based network}}
After getting the refined feature, we employ an attention-based network to predict the final importance score. Our attention-based network contains three components: i) a self-attention mechanism for modeling dependencies between any two frame feature; ii) a residual connection adding the output of self-attention mechanism and the original appearance feature, followed by layer normalization; iii) a linear layer and a sigmoid activation for predicting the final importance score.

The self-attention mechanism is to model the dependence between a desired output and every input. In this paper, we regard the appearance feature $X$ as query and the refined feature $Y$ as value and key. In other words, the appearance feature $X$ is projected as the query while the intermediate feature $Y$ is projected as key-value pair, i.e. 
\begin{gather}
{Q=X{W^Q}}\in{\mathbb R}^{F \times d}\\
{K=Y{W^K}}\in{\mathbb R}^{F \times d}\\
{V=Y{W^V}}\in{\mathbb R}^{F \times d}
\end{gather}
where $W^Q\in{\mathbb R}^{d\times d}$, $W^K\in{\mathbb R}^{d\times d}$ and $W^V\in{\mathbb R}^{d\times d}$ are the weights to be learned. After getting these values, we computes the dot products of the query with all keys, then divide by $\sqrt d$. Further, a soft-max function is applied to obtain normalized weights that reflect the correlations between frames. The results of soft-max function are then multiplied with the value $V$. The overall process can be denoted as:
\begin{equation}
H=\mathrm{Atten}\left(Q,K,V \right)=\mathrm{softmax}\left({\frac{{Q}{K^T}}{\sqrt{d}}}\right)V
\end{equation}

Like original transformer architecture, a layer normalization is employed for $H$. Finally, a linear layer by a sigmoid activation is used to predict the final frame score:
\begin{equation}
S=\mathrm{sigmoid}\left(\mathrm{linear}\left( \mathrm{norm}\left(H\right)\right)\right)
\end{equation}
where $S \in \mathbb R ^F$ are the importance scores of all frames in a video, $\mathrm{linear}$ denotes tow linear layers with 1024 and 1 neurons respectively.
\subsection{Discriminator}
To train the generator without any ground-truth labels, we design a discriminator to provide the needed optimization signal. Like typical GANs, our discriminator also aims to distinguish the original features $X$ from importance score weighted feature $\widetilde X$. Specifically, our discriminator is a LSTM layer with 1024 hidden units followed by a fully connected layer and a sigmoid activation function. The last hidden state of the LSTM layer is input to the fully connected layer. And the sigmoid activation function produces a binary-classification output. In actual, this label should be 1 for raw feature $X$ and 0 for score weighted feature $\widetilde X$. Once converged, the weighted feature will be  highly similar to raw frame feature, which means the main semantic information of an original video will be retained in the summary.

Note that the discriminator is only used when training the model. In testing, we will abandon the discriminator and only use the generator. In detail, the generator takes the appearance features $X$ extracted from a video as input and produces the importance score for every frame of the original video, which are then converted to key shots.

\subsection{Learning}
To train the model parameters, our learning objective includes three parts, whose specific form can be denoted as:
\begin{equation}
\mathcal{L}_{\mathrm{final}}=\mathcal{L}_{\mathrm{adv}}+\mathcal{L}_{\mathrm{rec}}+\mathcal{L}_{\mathrm{spar}}
\end{equation}
where $\mathcal{L}_{\mathrm{adv}},\mathcal{L}_{\mathrm{rec}},\mathcal{L}_{\mathrm{spar}}$ are the adversarial loss, reconstruction loss, and sparsity loss respectively. Note that we do not use any balancing coefficients. By minimizing this loss function, the parameters of our method can be learned. 

\subsubsection{Adversarial Loss}
The final goal of our method is to make the weighted frame features similar to original feature through adversarial training. As a result, we adopt the adversarial loss commonly used in GANs:
\begin{equation}
\begin{aligned}
\min \limits_{G}\max \limits_{D}\mathcal{L}_{\mathrm{adv}}\left(G,D\right)=&{\mathbb E}_X\left[\mathrm{log}D\left(X\right)\right]+{\mathbb E}_X\left[\mathrm{log}\left(1-D\left(G\left(X\right)\right)\right)\right]
\end{aligned}
\end{equation}
where $G\left(X\right)=\widetilde X$ is the output of the generator and $D\left(\cdot\right)$ denotes the binary classification output of the discriminator. Like traditional GANs, a minimax game occurs between the generator G and the discriminator D, where D is trained to maximize the probability of correct sample classification and G is simultaneously trained to minimize $\mathrm{log}\left(1-D\left(\widetilde X\right)\right)$. 

% This loss aims to match the distribution of weighted sequences produced by the generator with the data distribution of original sequences. 
\subsubsection{Reconstruction Loss}
An excellent summary should keep the main information of an original video. To achieve this goal, a reconstruction loss is introduced to minimize the information difference between the generated feature sequences $\widetilde X$ and the original feature sequences $X$. Specifically, we define the reconstruction loss $\mathcal{L}_{\mathrm{rec}}$ based on the output of the last hidden layer of LSTM in discriminator, since it captures the whole information of the input sequence. If we use $\phi \left(X\right)$ and $\phi \left(\widetilde X\right)$ to represent the output of the last hidden layer for input $X$ and $\widetilde X$ respectively, the reconstruction loss can be denoted as:
\begin{equation}
\mathcal{L}_{\mathrm{rec}}\left(X,\widetilde X\right)=\Vert{\phi \left(X\right)-\phi \left(\widetilde X\right)}\Vert_2
\end{equation}
where $\Vert \cdot \Vert_2$ is the Euclidean distance.

\subsubsection{Sparsity Loss}
A good summary should be short while keeping the important information. Therefore, we use the sparsity loss in \cite{Mahasseni2017} to make the bigger score value keep sparse. A high sparsity ratio gives a shorter and more concise summary. The detail formula is as following
\begin{equation}\label{Loss_spar}
\mathcal{L}_{\mathrm{spar}}\left(S\right)=\vert{\frac{1}{F}\sum_{f=1}^F s_f-\alpha}\vert
\end{equation}
where $\alpha$ is the percentage of the video's length to be preserved in the produced summary. We directly use 0.3 for $\alpha$ as suggested in \cite{Mahasseni2017}. The Eq. \eqref{Loss_spar} can be regarded as a summary-length regularization which penalizes having a large number of key frames selected in the summary, 

\subsection{Frame Scores to Key-shot Summary}
Given a test video, we will apply the trained model to obtain the frame-level importance scores $S$ and then convert them to key-shots. Firstly, we use the KTS algorithm \cite{Potapov2014} to detect scene change points which can serve as a potential key-shot segment. Secondly, we compute the average of all frames' importance scores in a shot as the shot-level score. To select some more important key-shots, we maximize the total scores of the selected key-shots while limit their total length to be $15\%$ length of original video. The maximization step is basically a $0/1$ Knapsack problem.

\subsection{Supervised Extension}\label{sec_sup}
Our unsupervised model can be easily extended to a supervised version if the ground-truth frame-level importance scores are given. In detail, we introduce the Mean Square Error (MSE) to minimize the distance between our predicted frame-level importance scores $S$ and the ground-truth scores $S'={\left\{s'_f\in{\left[0,1\right]}\right\}}_{f=1}^F$. This loss $\mathcal{L}_{\mathrm{sup}}$ can be denoted as:
\begin{equation}
\mathcal{L}_{\mathrm{sup}}=\mathrm{MSELoss}\left(S,S'\right)={\frac{1}{F}\sum_{f=1}^F \left(s_f-s'_f\right)^2}
\end{equation}

Then, we add this supervised loss $\mathcal{L}_{\mathrm{sup}}$ into the final loss. Mahasseni et al. \cite{Mahasseni2017} suggested replacing the sparsity loss with the supervised loss, but our experiments show that using all four kinds of loss can achieve better results. In other words, the final loss function can be written as: 
\begin{equation}
\begin{aligned}
\mathcal{L}_{\mathrm{sup\_final}}=\mathcal{L}_{\mathrm{adv}}+\mathcal{L}_{\mathrm{rec}}+\mathcal{L}_{\mathrm{spar}}+\mathcal{L}_{\mathrm{sup}}
\end{aligned}
\end{equation}
Note that we only change the loss function for supervised learning. Other things are kept the same. 

\section{Experiments}
This section first introduces the datasets and evaluation metrics. Then we compare the proposed model with some other state-of-the-art methods. Finally, we give the ablation analysis and some sample results.

\subsection{Experimental Settings}
\subsubsection{\textbf{Datasets}}
Our experiments mainly focus on the two popular benchmark datasets: SumMe \cite{Gygli2014} and TVSum \cite{Yale2015}. Some new datasets \cite{coview} are proposed recently, but we still use these two datasets for fair comparison. The SumMe dataset is a collection of 25 videos that cover a variety of events. The videos in SumMe are 1.5 to 6.5 minutes in length and most are first-person and third-person. The TVSum dataset contains 50 YouTube videos of 10 different categories (e.g. making sandwich, dog show, changing vehicle tire, etc.). The videos in this dataset are typically 1 to 5 minutes in length. Both of these two datasets have 15-20 user annotations. Since these two datasets are a little small, we also use the YouTube dataset \cite{Avila2011} and the Open Video Project (OVP) dataset \cite{OVP} to augment the training data. The YouTube dataset contains 39 videos that cover a variety of events including news, sports and cartoon while the OVP dataset contains 50 videos in different categories, such as documentary. Since different datasets provide the ground-truth annotations in various formats, we follow \cite{KeZhang2016} to generate the single set of ground-truth key-frames for each video in the datasets.

There exists two different methods for splitting a dataset into train and test part. One is the randomly splits, which randomly splits the dataset to $80\%$ videos for training and $20\%$ videos for testing. The final F-score is the average of multiple trials. According to the number of random splits, these test methods can be called as 5 Random, 10 Random, and Multiple Random. Although many previous methods employ this split, it may cause overlap in testing videos of different trials. To address this problem, He et al. \cite{He2019} proposed the 5-fold cross validation (5FCV) test method, which can evaluate the model on all videos in a dataset. Since the results with 5FCV are more reliable and reproducible, we will use the standard 5FCV as our default test method in the following discussion. 

To make a full comparison with other state-of-the-art methods, we follow \cite{KeZhang2016} to train and test our models on three different dataset settings: (1) Canonical: we calculate the F-score with 5FCV on each dataset individually. (2) Augmented: the training set is augmented with other three datasets in each fold of 5FCV. (3) Transfer: the model is trained using three datasets and tested on the remaining one.

\subsubsection{\textbf{Evaluation Metrics}}
Like other state-of-the-art methods \cite{KeZhang2016,Jiri2018}, we employ the F-score to evaluate the video summary results in most cases. Given the predicted summary $S$ and the ground truth summary $S'$, we first compute the precision $P$ and recall $R$ according to the temporal overlap between these two sets:
\begin{equation}
P=\frac{\mathrm{overlap\ duration\ between}\ S\ \mathrm{and}\ S'}{\mathrm{duration\ of}\ S}
\end{equation}
\begin{equation}
R=\frac{\mathrm{overlap\ duration\ between}\ S\ \mathrm{and}\ S'}{\mathrm{duration\ of}\ S'}
\end{equation}

Then, the final harmonic mean F-score can be obtained by
\begin{equation}
F=\frac{2P\times R}{P+R}\times 100\%
\end{equation}

Although the F-score is widely used, Mayu et al. \cite{Mayu2019} claimed that a random method can achieve comparable performance because of the well-designed pre-processing and post-processing. To remove these effect, they proposed to use rank correlation coefficients, i.e. Kendall’s $\tau$ and Spearman’s $\rho$, to measure the similarities between the rankings according to the predicted scores and human annotated. Refer to \cite{Mayu2019} for more details. Therefore, we will also employ this metric in some cases.

\subsubsection{\textbf{Implementation Details}}
Like other methods, we first downsample each video to 2 fps, and then utilize the output of the penultimate layer (pool5) of GoogleNet \cite{CSzegedy2015} pretrained on ImageNet as the frame representation, whose dimension is 1024. In other words, $d$ is set to 1024. The model is implemented with the PyTorch 1.3 framework. For canonical setting, we use the Adam optimizer with a learning rate of $3\times10^{-5}$ for the generator network. Another Adam optimizer with a learning rate of $1\times10^{-5}$ is used for the discriminator network. All experiments are conducted on an NVIDIA GTX 2080Ti GPU with 11GB memory.
%Instead of minimizing the feature distance, $\mbox{SASUM}_{unsup}$ \cite{Hua2018} extract the semantically relevant video segments via minimizing the distance between the generated description sentence of the summarized video and the human annotated text of the original video.

\begin{table*}[ht]
	\centering
		\caption{Comparison of the proposed method with other unsupervised approaches in canonical, augmented and transfer settings on the SumMe and TVSum datasets in terms of harmonic F-score.}
	\label{Table_unsup}
	\renewcommand\arraystretch{1.2}
	{
	\begin{tabular}{c c c c c c c c}  
		\toprule
		\multirow{2}*{Method}& \multicolumn{3}{c }{SumMe}& \multicolumn{3}{c}{TVSum}& \multirow{2}*{Test. method} \\
		\cline{2-7}
		& Can. &Aug. &Tr. &Can &Aug. &Tr. \\
		\hline
		%$\mbox{Video-MMR}\cite{YLi2011}$  & 26.6 & - & -& - &- &-&-\\
		
		$\mbox{SUM-GAN}_{dpp}$\cite{Mahasseni2017} &39.1&43.4&-&51.7&59.5&-&5 Random\\
		
		$\mbox{DR-DSN}$ \cite{ZhouKaiyang2017} &41.4&42.8&42.4&57.6&58.4&57.8&5FCV\\

        %$\mbox{SASUM}_{unsup}\cite{Wei2018}$ &40.6&-&-&53.9&-&-&Multiple Random\\
		
		$\mbox{SUM-FCN}_{unsup}$\cite{Mrigank2018} &41.5&-&-&52.7&-&-&Multiple Random\\

        $\mbox{Cycle-SUM}$\cite{LYuan2019} &41.9&-&-&57.6&-&-&5 Random\\
		
		$\mbox{ACGAN}$ \cite{He2019} &46.0&47.0&44.5&58.5&58.9&57.8&5FCV\\
		
		$\mbox{SUM-GDA}$ \cite{li2020exploring} &50.0&50.2&46.3&\textbf{59.6}&{\bf 60.5}&{\bf 58.8}&5FCV\\
		
		$\mbox{GL-RPE}$ \cite{jungglobal} &50.2& - & - &59.1&-&-&5FCV\\
		\midrule
		$\mbox{CAAN(ours)}$ &{\bf 50.8}&{\bf 50.9}& {\bf46.5}&{\bf59.6}&59.8&57.8&5FCV\\		
		\bottomrule
	\end{tabular}}
\end{table*}

\subsection{Quantitative Results}
\subsubsection{\textbf{Comparison with some unsupervised approaches}}
In this section, we compare our model with several state-of-the-art methods for unsupervised video summarization. Most of them are GAN-based approaches, including $\mbox{SUM-GAN}_{dpp}$, $\mbox{ACGAN}$, $\mbox{Cycle-SUM}$ and $\mbox{GL-PRE}$. $\mbox{SUM-GAN}_{dpp}$ \cite{Mahasseni2017} employed three LSTM to implement the overall GAN framework. Inspired by the success of cycle-GAN architecture in image generation, $\mbox{Cycle-SUM}$ \cite{LYuan2019} proposed the  Cycle-consistent Adversarial LSTM architecture, which can effectively reduce the information loss in the summary video. $\mbox{ACGAN}$ \cite{He2019} utilized a conditional feature selector to guide GAN model, which can make the model focus on more important temporal regions of the whole video frames. $\mbox{GL-PRE}$ \cite{jungglobal} further combined the relative position embedding with global and local decomposition strategy. Besides this major kind, some methods use some heuristic criterion to guide the learning. Based on deep reinforcement learning, $\mbox{DR-DSN}$ \cite{ZhouKaiyang2017} proposed to jointly explain the diversity and representativeness of the generated summaries. $\mbox{SUM-FCN}_{unsup}$ \cite{Mrigank2018} augmented the reconstruction loss with a diversity regularization to optimize the parameters in their fully convolutional sequence network. $\mbox{SUM-GDA}$ \cite{li2020exploring} considers pairwise temporal relations of video frames by modeling the relations within paired frames as well as the relations among all pairs.

Table \ref{Table_unsup} shows the results of our model against the above unsupervised methods.  On the two benchmarks, the proposed model outperforms our baseline under any settings. For example, on canonical setting, the F1 score of our method is $11.5\%$ and $6.1\%$ higher that that of the $\mbox{SUM-GAN}_{dpp}$ on two benchmark datasets. And compared with the ACGAN, which also contains self-attention and GAN, our method obtains better performance. And the improvement is larger than $4\%$ and $1\%$ on the canonical setting. Finally, our proposal achieves the best performance on the SumMe dataset and is comparable to the state-of-the-art $\mbox{SUM-GDA}$ on the TVSum dataset. We think the main reasons was that $\mbox{SUM-GDA}$ uses repelling loss to enhance the diversity among video frames. In the future, we will explore to integrate this loss to our method.

\begin{table}[t]
	\caption{Comparison of our supervised version with other supervised approaches in canonical setting on the SumMe and TVSum datasets in terms of harmonic F-score.}
	\label{Table_sup}
	\centering
	\renewcommand\arraystretch{1.2}
	\begin{tabular}{c c c c}  
		\toprule
		Method&SumMe&TVSum&Testing method\\
		\hline
       $\mbox{DPP-LSTM}$ \cite{KeZhang2016} &38.6 &54.7 &-\\
	   $\mbox{SUM-GAN}_{sup}$ \cite{Mahasseni2017}  &41.7 &56.3 &5 Random\\
	   $\mbox{DR-DSN}_{sup}$ \cite{ZhouKaiyang2017} &42.1&58.1&5FCV\\
       $\mbox{SASUM}_{sup}$ \cite{wei2018} &45.3 &58.2 &Multiple Random\\
		${SUM-FCN}$ \cite{Mrigank2018} &47.5 &56.8 &Multiple Random\\
		$\mbox{M-AVS}$ \cite{Ji2020video} &44.4 & 61.0 &5 Random\\
        $\mbox{Cycle-SUM}_{sup}$\cite{LYuan2019} &44.8&58.1&5 Random\\
		$\mbox{ACGAN}_{sup}$ \cite{He2019} &47.2 &59.4& 5FCV\\
		$\mbox{Stack-LSTM}$ \cite{wang2019stacked} &49.2 &60.8& 5FCV\\
		$\mbox{SMN}$ \cite{wang2019stacked} &\textbf{58.3} &\textbf{64.5} & 5FCV\\
		\midrule
		$\mbox{CAAN}_{sup}$ & 50.6 & 59.3&5FCV\\		
		\bottomrule
	\end{tabular}
\end{table}

\subsubsection{\textbf{Comparison with some supervised approaches}}
As stated in the section \ref{sec_sup}, our model can be easily extended to a supervised version. Here, we compare this supervised version with some other state-of-the-art supervised methods. For simplicity, we just use the canonical setting. The F-scores on the SumMe and TVSum datasets are illustrated in Table \ref{Table_sup}. On these datasets, our supervised model outperforms most of the supervised methods. In particular, we achieve the second best F-score on the SumMe dataset and the gain is larger than 1.4\%. On the TVSum dataset, our performance is a little lower, but it is still comparable to many methods except the SMN. Our intuition is that these methods mainly obtain great performance improvement only on one dataset while our method can achieve quite good performance on both datasets. According to \cite{wang2019stacked}, the large performance improvement of $\mbox{SMN}$ mainly depends on the memory layers, which can be combined with our method in the future. In a word, this comparison study demonstrates that our supervised version still obtain excellent performance. Moreover, we can find that our unsupervised model performs better than many supervised models on the SumMe dataset by comparing the F-scores in Table \ref{Table_unsup} and \ref{Table_sup}. This further shows the advantage of our unsupervised method.

\begin{table}[t]
	\caption{Comparison of our unsupervised method with other approaches on the TVSum dataset in terms of Kendall’s $\tau$ and Spearman’s $\rho$ correlation coefficients.}
	\label{Table_correlation}
	\centering
	\renewcommand\arraystretch{1.2}
	\begin{tabular}{c c c}  
		\toprule
		Method & Kendall’s $\tau$ & Spearman’s $\rho$ \\
		\hline
		$\mbox{Random}$ &0.000&0.000\\
		$\mbox{DR-DSN}$\cite{ZhouKaiyang2017} &0.020&0.026\\
		$\mbox{DPP-LSTM}$\cite{KeZhang2016}  &0.042&0.055\\
		$\mbox{CAAN(ours)}$ &{\bf 0.062}&{\bf 0.090}\\
		\midrule
		$\mbox{Human}$ &0.177&0.204\\		
		\bottomrule
	\end{tabular}
\end{table}

Finally, we evaluate the performance using the rank order statistics, which measure the similarities between the implicit rankings provided by generated and human annotated frame level importance scores. Like Mayu et al. \cite{Mayu2019}, we only report the scores of TVSum dataset. The results of our method with two video summarization methods are illustrated in Table \ref{Table_correlation}. For a reference, we also add human performance and the results using random scoring. Compared with these methods, we achieve the biggest correlation coefficients on both Kendall’s $\tau$ and Spearman’s $\rho$ except the human.

\subsection{Ablation Analysis}
In this section, we conduct various ablation analysis to evaluate the impact of different components and loss settings of our unsupervised CAAN model. For simplicity, we only adopt canonical evaluation setting and the F-score metric.

\begin{table}[t]
	\caption{Performance (F-score, $\%$) of our framework with different components on the SumMe and TVSum datasets.}
	\label{Table_ablation_compon}
	\centering
	\renewcommand\arraystretch{1.2}
	\begin{tabular}{c c c}  
		\toprule
		Method&SumMe&TVSum\\
		\hline
		$\mbox{CAAN}_{w/o-F}$  &45.06 &57.86\\
		$\mbox{CAAN}_{w/o-A}$ &45.03&55.17\\
		$\mbox{CAAN}_{Q=Y}$ &48.55&57.01\\
		\midrule
		$\mbox{CAAN}$ &{\bf 50.81}&{\bf 59.58}\\		
		\bottomrule
	\end{tabular}
\end{table}

\subsubsection{\textbf{The impact of different components}} As stated in previous sections, our generator consists of two parts: the FSCN and the attention based part. To investigate their effect, we conduct the experiments with two model variants, which are $\mbox{CAAN}_{w/o-F}$ and $\mbox{CAAN}_{w/o-A}$ respectively. In the former, we remove the FSCN and only use the attention-based network, whose query, key and value are projected from the same appearance feature $X$. In contrast, $\mbox{CAAN}_{w/o-A}$ drops the attention-based network part. To predict the final importance score, we add a linear layer and a sigmoid activation after the FSCN. 

The results on SumMe and TVSum are shown in Table \ref{Table_ablation_compon}. Compared the F-scores between $\mbox{CAAN}_{w/o-F}$ and $\mbox{CAAN}$, we can see the global representation can largely improve the performance on both two datasets. The reason may be that the global representation already contains a certain level of long-range dependencies, which can help our model focus on the temporal information more accurately. The performance of $\mbox{CAAN}_{w/o-A}$ on two datsets are $45.03\%$ and $55.17\%$ respectively, which is $5.78\%$ and $4.41\%$ lower than the proposed $\mbox{CAAN}$ architecture. This demonstrates the effectiveness of this attention part. Besides, the basic architecture of variant $\mbox{CAAN}_{w/o-A}$ is very similar to $\mbox{SUM-FCN}_{unsup}$ in \cite{Mrigank2018}. However, our GAN-based training method and more skip connections can increase the performance from 41.5\% and 52.7\% to 45.03\% and 55.17\%.

As illustrated in Fig. \ref{Fig_frame}, the self-attention unit needs a query $Q$ and a set of key-value pairs $\left(K,V\right)$. And we project the original feature $X$ as the query. Therefore, it is interesting to explore the effect of using different query on the final performance. We use $\mbox{CAAN}_{Q=Y}$ to denote the model variant of initializing query $Q$ with the intermediate feature $Y$ instead of the appearance feature $X$. Its results on two datasets are given in the fourth line of Table \ref{Table_ablation_compon}. We can see that the performance of this variant decreases by $2.26\%$ and $2.57\%$ on SumMe and TVSum. This comparison shows that the model can get more complete information about the entire video when using the appearance feature $X$ to initialize $Q$.

%We analyze how the performance of our model is related to the global representation. The global representation in our model extracts the rich temporal information of the entire input video. It is fed into the Attention-based network together with the original appearance feature. In TABLE \ref{Table_ablation_compon}, $\mbox{CAAN}_{w/o-F}$ demonstrates a case study of the final importance scores generated without the global representation. We can see that the global representation improves the performance on both datasets. The reason may be that the global representation already contains a certain level of long-range dependencies and it can help our model focuses more accurately on the temporal information. 

%We also test the performance of a model variant $\mbox{CAAN}_{w/o-A}$ that drop the Attention-based network part, and use a linear layer and a sigmoid activation for predicting the final frame score behind the Full Convolutional Sequence network in order to analyze the effect of self-attention in the proposed model. From Table \ref{Table_ablation_compon}, we can see that the performance of this variant is only $45.01\%$ on SumMe, which is lower than most other ablation models and is $5.5\%$ lower than the proposed $\mbox{CAAN}$ architecture. This demonstrates that the Self-attention mechanism helps learn better video temporal representation resulting in more compact and complete summaries.

\subsubsection{\textbf{The influence of different loss settings}}
The final loss of our unsupervised model is composed of adversarial loss, reconstruction loss and sparsity loss. To verify their impact, we have conducted experiments by removing one of them while other things are kept the same. The experimental results are shown in Table \ref{Table_ablation_loss}. In this table, the $\mbox{CAAN}_{w/o-spar}$ refers the model without using sparsity regularization while the $\mbox{CAAN}_{w/o-rec}$ denotes the model without using reconstruction loss. From this table, we can see that removing any part will damage the F-score. Specifically, the F-score without sparsity loss decreases by 2.20\% and 1.79\% respectively compared with the original model. These results indicate that the sparsity loss provides more regularization and ensures that no excessive key frames are produced as summaries. When dropping the the reconstruction loss, the F-score becomes much worse and is even lower than $\mbox{CAAN}_{w/o-spar}$, which means that the reconstruction loss is more important. The reason is that the reconstruction loss can make the generated feature sequences similar to the original feature sequences of the input video. In a word, both the reconstruction loss and the sparsity loss are important for our model.

\begin{table}[t]
	\caption{Performance (F-score, $\%$) of our framework using different loss settings on the SumMe and TVSum datasets.} \label{Table_ablation_loss}
	\centering
	\renewcommand\arraystretch{1.2}
	\begin{tabular}{c c c}  
		\toprule
		Method&SumMe&TVSum\\
		\hline
		$\mbox{CAAN}_{w/o-spar}$  &48.61 & 57.79\\
		$\mbox{CAAN}_{w/o-rec}$ &48.53 &56.84\\
		\midrule
		$\mbox{CAAN}$ &{\bf 50.81}&{\bf 59.58}\\		
		\bottomrule
	\end{tabular}
\end{table}

\begin{figure}[t!]
	\centering
	\includegraphics[width=0.85\columnwidth]{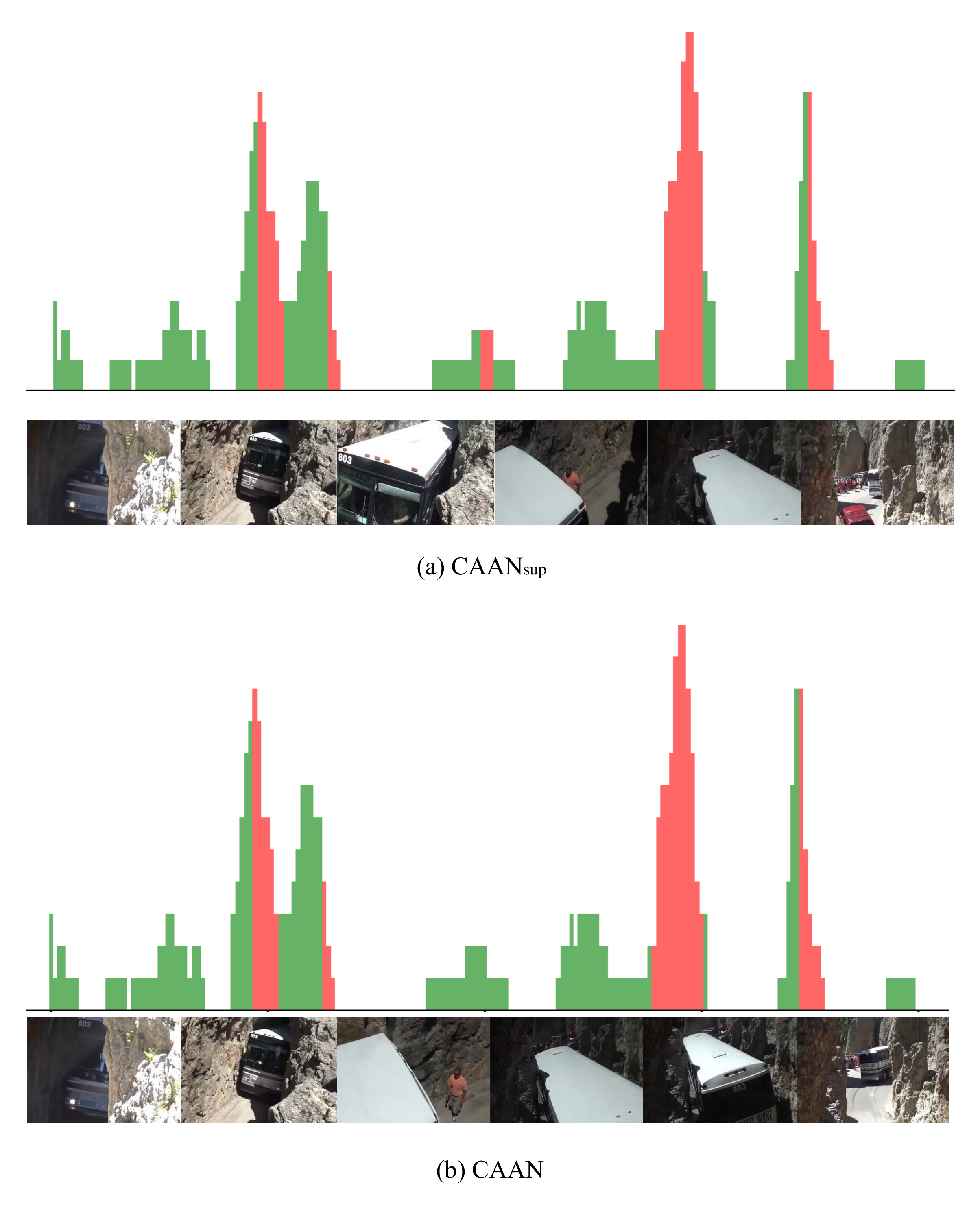}\\
	\caption{An exemplar video summarization result of our supervised version $\mbox{CAAN}_{sup}$ and the overall model $\mbox{CAAN}$ for the test video {\em bus in rock tunnel} in the SumMe dataset. The green bars and red regions in (a) and (b) are the ground-truth frame-level importance scores and the selected frames respectively. Some sample frames in predicted summary are also showed.}
	\label{Fig_qual}
\end{figure}

\subsection{Qualitative results}
To better demonstrate the effectiveness of our framework, we provide an exemplar video summarization result by the supervised version $\mbox{CAAN}_{sup}$ and the overall unsupervised model $\mbox{CAAN}$ in Fig. \ref{Fig_qual}. The input is a video in the SumMe dataset, where a bus is driving in rock tunnel. The green bars in Fig. \ref{Fig_qual} denote the ground-truth frame-level importance scores while the red regions are the selected subsets by different methods. Some selected frames are shown in the bottom of the figure.

In general, both of our methods produce high-quality summaries that capture most of the peaks in ground-truth importance score. The generated summary by unsupervised model covers the timeline, while the summary produced by supervised learning is more complete with story-line. This is because $\mbox{CAAN}_{sup}$ uses extra human annotated labels, which encourages the produced summary closer to human subjective consciousness.

\section{Conclusion}
This paper presents a convolutional attentive adversarial network for unsupervised video summarization, which consists of a generator and a discriminator. The generator predicts frame-level importance scores while the discriminator tries to distinguish the score-weighted frame features and the original frame features. Moreover, the generator integrates a fully convolutional sequence network and attention based network, which models global relation and capture the long-range temporal dependence of a video respectively. Once converged, the generator can produce the summary, which keeps the important content of original video. Without any recurrent structure, the model is easy to be paralleled, which can reduce the running time largely. Extensive experiments conducted on two benchmark datasets show the superiority of our method against other state-of-the-art unsupervised approaches, and even outperforms some published supervised approaches.

In the future, we will augment the model with extra memory layer to further improve the performance and test the method in large-scale dataset for video summarization. Besides, we will also explore the proposed architecture in other areas, such as video caption, image text recognition.  

\section*{Acknowledgements}
\label{}
This work was supported in part by the National Natural Science Foundation of China (No.61902321, No. U19B2037), the Ministry of Science and Technology Foundation funded project under Grant 2020AAA0106900, the China Postdoctoral Science Foundation funded project under Grant 2019M653746.

\bibliographystyle{elsarticle-num} 
\bibliography{IEEEabrv,refer}

\end{document}